\documentclass[letterpaper, 10 pt, conference]{ieeeconf}  
\pdfoutput=1

\IEEEoverridecommandlockouts                              

\usepackage{graphics} 
\usepackage{subfigure}
\usepackage{epsfig} 
\usepackage{amsmath} 
\usepackage[T1]{fontenc}
\usepackage{bm}
\usepackage{cite}
\usepackage{booktabs}
\usepackage{multirow}
\usepackage{diagbox}
\usepackage{threeparttable}
\usepackage{colortbl}
\usepackage{hyperref}
\hypersetup{
    colorlinks=false, 
    }
\hypersetup{hidelinks} 
\newcommand \transpose {\mathsf{T}} 
\newcommand \dlo {\rm dlo}
\newcommand \arm {\rm arm}

\title{\LARGE \bf
A Coarse-to-Fine Framework for Dual-Arm Manipulation of Deformable Linear Objects with Whole-Body Obstacle Avoidance
}

\author{Mingrui Yu, Kangchen Lv, Changhao Wang, Masayoshi Tomizuka, and Xiang Li
\thanks{M. Yu, K. Lv, and X. Li are with the Department of Automation, Tsinghua University, Beijing, China.
C. Wang and M. Tomizuka are with the Department of Mechanical Engineering, University of California, Berkeley, CA, USA.
This work was supported in part by the Science and Technology Innovation 2030-Key Project under Grant 2021ZD0201404, in part by the National Natural Science Foundation of China under Grant U21A20517 and 52075290, 
and in part by the Institute for Guo Qiang, Tsinghua University. Corresponding author: Xiang Li (xiangli@tsinghua.edu.cn)}}

\begin{document}

\maketitle
\thispagestyle{empty}
\pagestyle{empty}

\begin{abstract}
Manipulating deformable linear objects (DLOs) to achieve desired shapes in constrained environments with obstacles is a meaningful but challenging task. Global planning is necessary for such a highly-constrained task; however, accurate models of DLOs required by planners are difficult to obtain owing to their deformable nature, and the inevitable modeling errors significantly affect the planning results, probably resulting in task failure if the robot simply executes the planned path in an open-loop manner. In this paper, we propose a coarse-to-fine framework to combine global planning and local control for dual-arm manipulation of DLOs, capable of precisely achieving desired configurations and avoiding potential collisions between the DLO, robot, and obstacles. Specifically, the global planner refers to a simple yet effective DLO energy model and computes a coarse path to find a feasible solution efficiently; then the local controller follows that path as guidance and further shapes it with closed-loop feedback to compensate for the planning errors and improve the task accuracy. Both simulations and real-world experiments demonstrate that our framework can robustly achieve desired DLO configurations in constrained environments with imprecise DLO models, which may not be reliably achieved by only planning or control.
\end{abstract}

\section{Introduction}
Robotic manipulation methods for rigid objects have been extensively studied and reliably applied to manufacturing and services \cite{billard2019trends}.
In addition to rigid objects, deformable linear objects (DLOs) are also common in human life, such as cables, wires, ropes, rods, etc \cite{jose2018robotic}; however, there are new challenges coming from the deformable properties of DLOs when applying classical manipulation methods to DLO manipulation \cite{yin2021modeling,zhu2021challenges}. 

\begin{figure} [tb]
  \centering 
    \includegraphics[width=8.6cm]{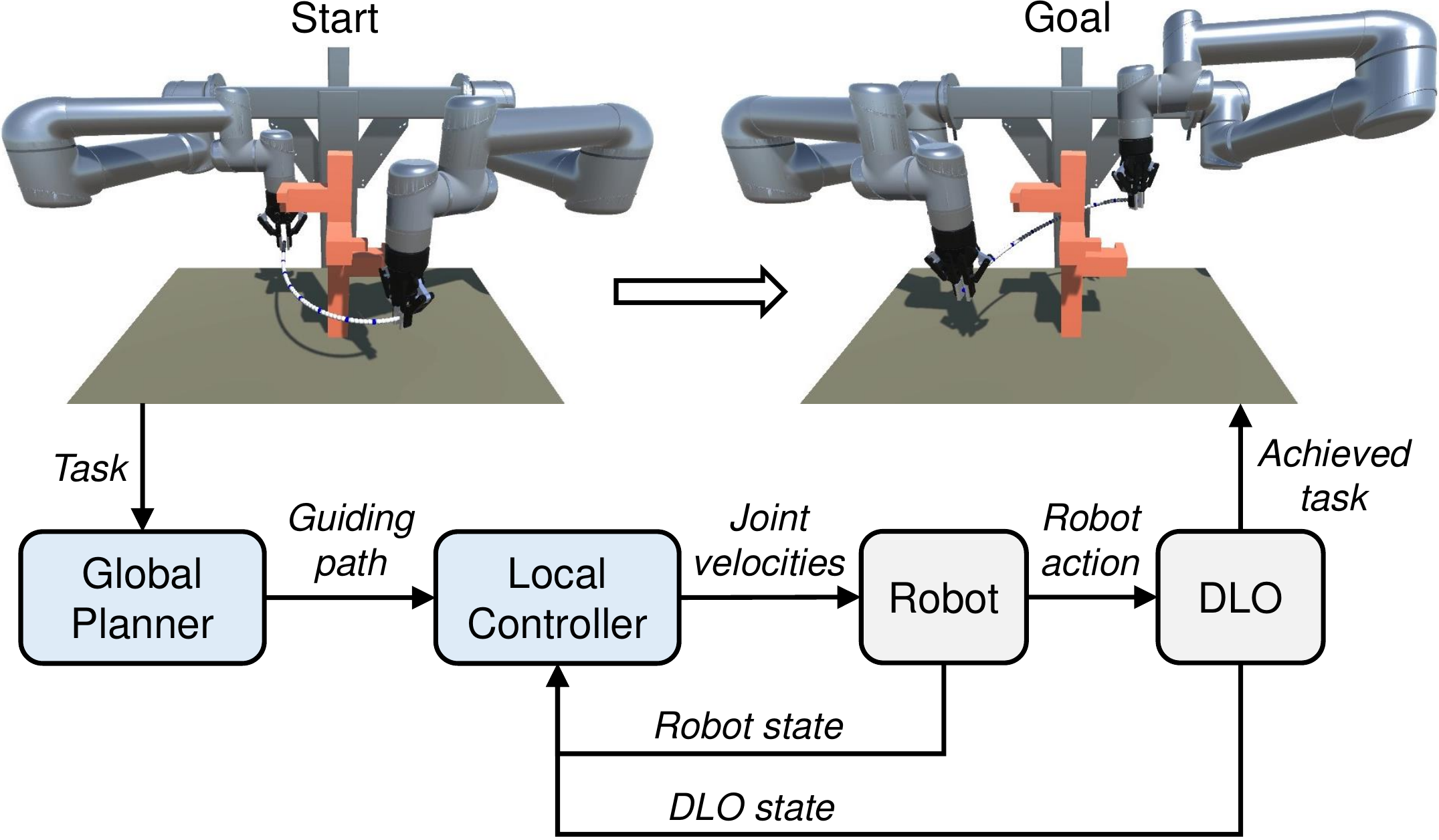} 
  \vspace{-3mm}
  \caption{Overview of the proposed framework for dual-arm manipulation of DLOs in constrained environments with obstacles, in which the global planning and local control complement each other to robustly achieve complex tasks even using imprecise DLO models. 
  The collision avoidance of both the DLO and robot body is considered in the planning and execution.}
  \label{fig:overview}
    \vspace{-2mm}
\end{figure}

We focus on a general problem of DLO manipulation: dual-arm manipulating a DLO from a start configuration to a desired (goal) configuration, which contains both moving and shaping of the DLO. 
In our previous work \cite{yu2022global,yu2021shape}, we have proposed an adaptive controller which can stably and efficiently achieve DLO shape control in unobstructed environments. 
In this work, we further deal with a more complex and practical scenario: 
manipulating DLOs in constrained environments with obstacles (see Fig. \ref{fig:overview}), which requires not only accurate final control results but also proper collision-free moving paths of both DLOs and robot bodies.

A series of works have studied the DLO shaping from the perspective of control, which use the real-time error between the current shape and desired shape as the feedback for closed-loop servoing \cite{navarro2016Automatic,wang2022offline,zhu2021vision,lagneau_automatic_2020,jin2019robust}. 
However, these control-only methods usually consider a simplified scenario: no obstacles exist, and the robot end-effectors can move freely without 
considering the arm bodies; 
moreover, local controllers cannot preventing DLOs from falling into local optimal shapes when desired deformations are large. 

To achieve DLO manipulation in constrained environments, global path planning is necessary. Existing studies have proposed different DLO models and incorporated them into classical high-dimensional path planning methods \cite{wakamatsu2004static,moll2006path,bretl2014quasi}, such as rapidly-exploring random trees (RRTs) \cite{lavalle1998rapidly} and probabilistic roadmap methods (PRMs) \cite{amato1996randomized}. 
Paths of DLO configurations from the start to the goal are planned, and corresponding end-effector trajectories are extracted for open-loop executions \cite{roussel2020motion,sintov2020motion,mitrano2021learning}. 
However, the planning-only methods are more affected by the inevitable DLO modeling errors than control, since no real-time feedback compensates for the errors. The planned end-effector trajectory may not move the DLO exactly along the expected configuration path, which may cause failures of the actual execution. 
As a result, most of these approaches are restricted to simulations with sufficiently accurate models. 



To overcome the respective shortcomings of the existing planning-only and control-only methods, we propose a new framework that combines both global planning and local control, with the structure shown in Fig. \ref{fig:overview}. The planner efficiently computes a coarse and global collision-free path to the desired configuration, based on a simple yet effective DLO model; Then, the controller tracks the planned path as guidance while using closed-loop feedback to compensate for the modeling errors in the planning phase and locally avoid obstacles.
In such a framework, the global planning and local control complement each other: the former achieves the finding of feasible solutions and the latter improves the accuracy of the task,
which hence enables the robot to robustly and precisely manipulating DLOs to desired configurations in constrained environments while avoiding potential collisions.
Specifically, for the planner, we use an RRT planning framework with a DLO energy model for projecting a random DLO configuration to a stable one; for the controller, we formulate it as a model predictive control (MPC) problem with artificial potentials, in which we apply our previously proposed Jacobian model learning method \cite{yu2022global}.
The full configurations of the DLO and robot are considered in the global planning and local control.
We carry out both simulations and real-world experiments to demonstrate that our method can achieve various DLO manipulation tasks in constrained environments which may not be reliably achieved by only planning or control. 
The video is available on the project website\footnote{Project website: \url{https://mingrui-yu.github.io/DLO_planning}}.

\section{Related Work}

\subsection{DLO Path Planning}
To use sampling-based planning methods for high-dimensional DLO path planning, an approach to generate valid DLO configurations is required.
Bretl and McCarthy \cite{bretl2014quasi} formulated the DLO static equilibrium as an optimal control solution and derived that the configuration space of an one-end-fixed Kirchhoff elastic rod is a six-dimensional manifold, but it requires to solve complicated differential equations to determine whether a sample is stable.
Roussel et al. \cite{roussel2020motion} used this model for global DLO motion planning in complex tasks at the cost of very huge computation time. To reduce the online time cost, \cite{sintov2020motion} pre-computed a roadmap of stable DLO configurations.
However, the roadmap takes a lot of time to compute and memory space to store, and the planning performance highly depends on the roadmap size. 
Furthermore, this ideal model cannot consider gravity, and \cite{mishani2022realtime} showed that it may only be appropriate for DLOs of very high stiffness, such as nitinol rods.

An alternative way is to compute stable DLO equilibrium configurations by locally minimizing the deformation energy using general optimization approaches \cite{wakamatsu2004static,moll2006path}.
We hold the view that such methods are more general and suitable for DLO path planning since we can use it to efficiently project a random invalid configuration to a valid one and incorporate it into standard single-query planning algorithms.
In this work, we employ a discrete DLO energy model and an RRT planning framework, achieving efficient and practical DLO path planning.

Recently, learning non-physical DLO models for planning from data using neural networks has also been studied. 
In \cite{mitrano2021learning}, three networks are trained for planning and execution, including a state-action forward prediction model, a classifier for predicting the validity of the model, and a recovery policy. 
The problems are that such methods need to pre-collect a lot of data of the manipulated DLO for training, and their effectiveness on different untrained DLOs and scenarios cannot be guaranteed.

\subsection{DLO Shape Control}
Controlling DLOs to desired shapes using local controllers has been widely researched \cite{navarro2016Automatic,wang2022offline,zhu2021vision,lagneau_automatic_2020,jin2019robust}, and related methods have been detailedly reviewed in \cite{yu2022global}. However, most of them cannot handle large deformations, and the manipulation processes are usually not smooth.

In our previous work \cite{yu2022global}, we proposed an efficient and adaptive approach to learning a state-action DLO model and achieved stable and smooth large deformation control, which makes it possible to track planned paths. 
In this work, we additionally apply the artificial potential method for local obstacle avoidance, since the actual execution path may not be exactly the planned collision-free path.

\subsection{Combining Planning and Control for DLO Manipulation}

McConachie et al. \hspace{-1mm} \cite{mcconachie2020manipulating} proposed another framework for interleaving planning and control for manipulating deformable objects. 
Their controller first attempts to perform tasks directly; if their deadlock predictor predicts the controller will get stuck, their planner will be invoked to move the object to a new region. 
Note that their aim and combination approach are different from ours: first, their planner uses a more simplified model called ``virtual elastic band" focusing on the overstretch of deformable objects caused by grippers or obstacles; second, their planned path of grippers is open-loop executed without local adjustment using real-time feedback;
as a result, the DLO may be over-compressed or hooked by obstacles.
In contrast, our method considers the full DLO configurations both in planning and control, and employs feedback control during the whole manipulation process.


\section{Preliminaries} \label{section:problem_formulation}
In this work, a dual-arm robot grasps the two ends of the DLO and quasi-statically manipulates the DLO to the desired configuration, during which both the DLO and the arms should not collide with obstacles. 
The configuration of the DLO is represented by the positions of multiple features uniformly distributed along the DLO. 

Some frequently-used notations are listed as follows. 
The position of the $k^{\rm th}$ feature is represented as $\bm x_k \in \Re^3$.  
The configuration vector of the DLO is represented as $\bm{x} := [\bm{x}_1; \cdots; \bm{x}_m] \in \Re^{3m}$ , where $m$ is the number of the features.
The joint position of the dual-arm robot is represented as $\bm q \in \Re^n$, where $n$ is the degrees of freedom (DoFs) of the robot. The configuration vector of the robot end-effectors is represented as $\bm r \in \Re^{12}$.

The control input $\bm u$ to the system is the joint velocity of the dual-arm robot. That is
\begin{equation}
   \dot{\bm q} =  \bm u
\end{equation}

\section{Global Planning}

The planner is first invoked to efficiently find a global collision-free path in constrained environments using a simple yet effective DLO energy model. This coarse path will be sent to the controller as global guidance, to achieve the solving of the complex manipulation task.


\subsection{Projection to Stable DLO Configuration Manifold}
As defined in Section \ref{section:problem_formulation}, the raw configuration space of the DLO is $3m$ dimensional; however, only a subspace contains stable equilibrium configuration, which can be called a manifold. Thus, the planning is constrained on this manifold, i.e., the DLO configuration of each node should be guaranteed stable. 
Finding a stable configuration by randomly sampling in the raw space with rejection strategies is unlikely because the dimension of the stable space is much less than that of the raw space. Alternatively, it is more appropriate to use projection methods to move a random sample in the raw space onto its neighbor constrained manifold \cite{berenson2009manipulation}.

Such projections can be achieved by utilizing a DLO energy model. Denote the potential energy of an elastic DLO as $E$, which is assumed to be fully determined by the DLO configuration $\bm x$. A stable equilibrium of the DLO with two end poses fixed is where the DLO's internal configuration locally minimizes the potential energy $E$ \cite{wakamatsu2004static,moll2006path}. 
That is
\begin{equation}
    \frac{\partial E}{\partial \bm x_k} = \bm 0, \quad \forall k = 3, \cdots, m-2
\end{equation}
where the first two and last two feature points are fixed to represent the fixed end poses in discrete DLO models \cite{bergou2008discrete}.

A random sample in the raw space $\bm x^0$ can be projected onto the stable configuration manifold by 
formulating it as a local minimization problem of the energy
with $\bm x^0$ as the initial value:
\begin{equation} \label{eq:energy_optimize}
\begin{aligned}
    \bm x^{\rm stable} = 
     \arg  \min_{\bm x} 
     & \quad   E(\bm x)
    \\
    \text{s.t.} 
    \quad  & \bm x_k = \bm x_k^0, \quad \forall k = 1, 2, m-1, m
\end{aligned}
\end{equation}
More constraints may be incorporated for specific tasks. We denote the process of projecting a raw configuration to a neighbor stable configuration as $\bm x^{\rm stable} = \text{ProjectStableConfig}(\bm x^{0})$.

\begin{figure} [tb]
  \centering 
  \subfigure[]{ 
    \label{fig:mass_spring_model} 
    \includegraphics[width=3.7cm]{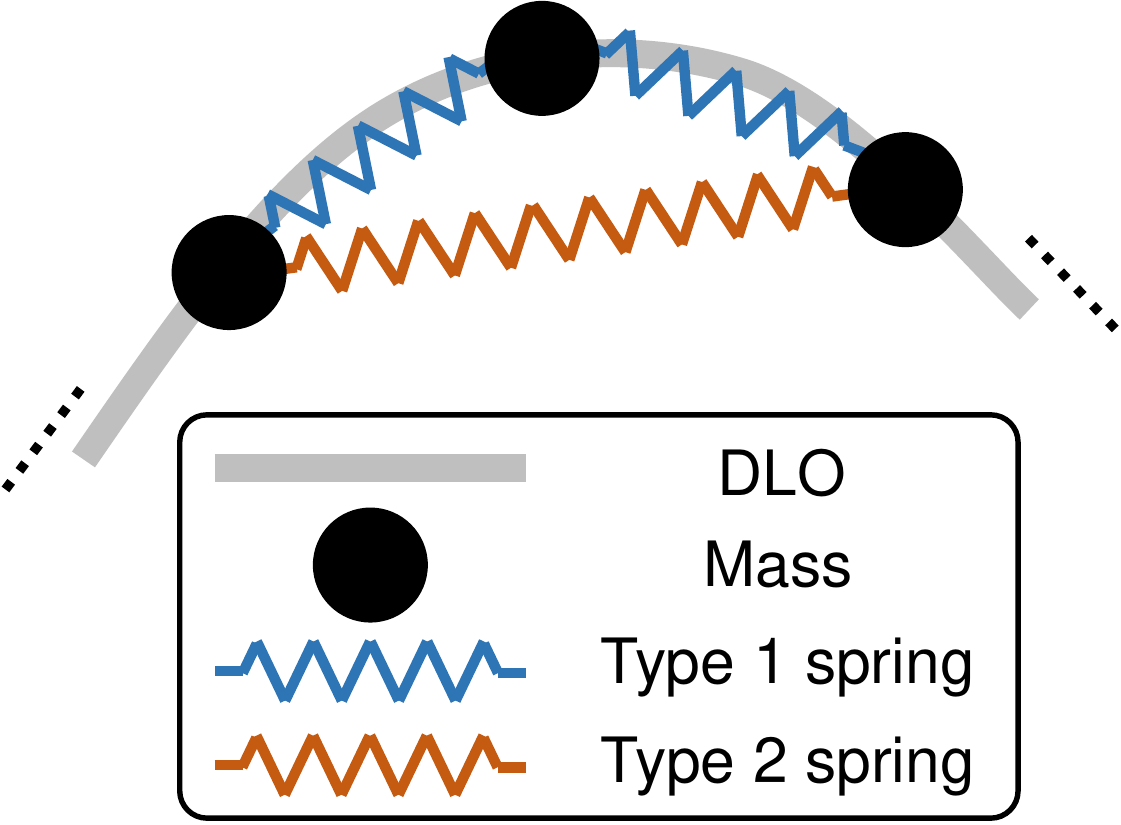} 
  } 
  \hspace{-0.6cm}
  \subfigure[]{ 
    \label{fig:projection_example_1} 
    \includegraphics[width=2.5cm]{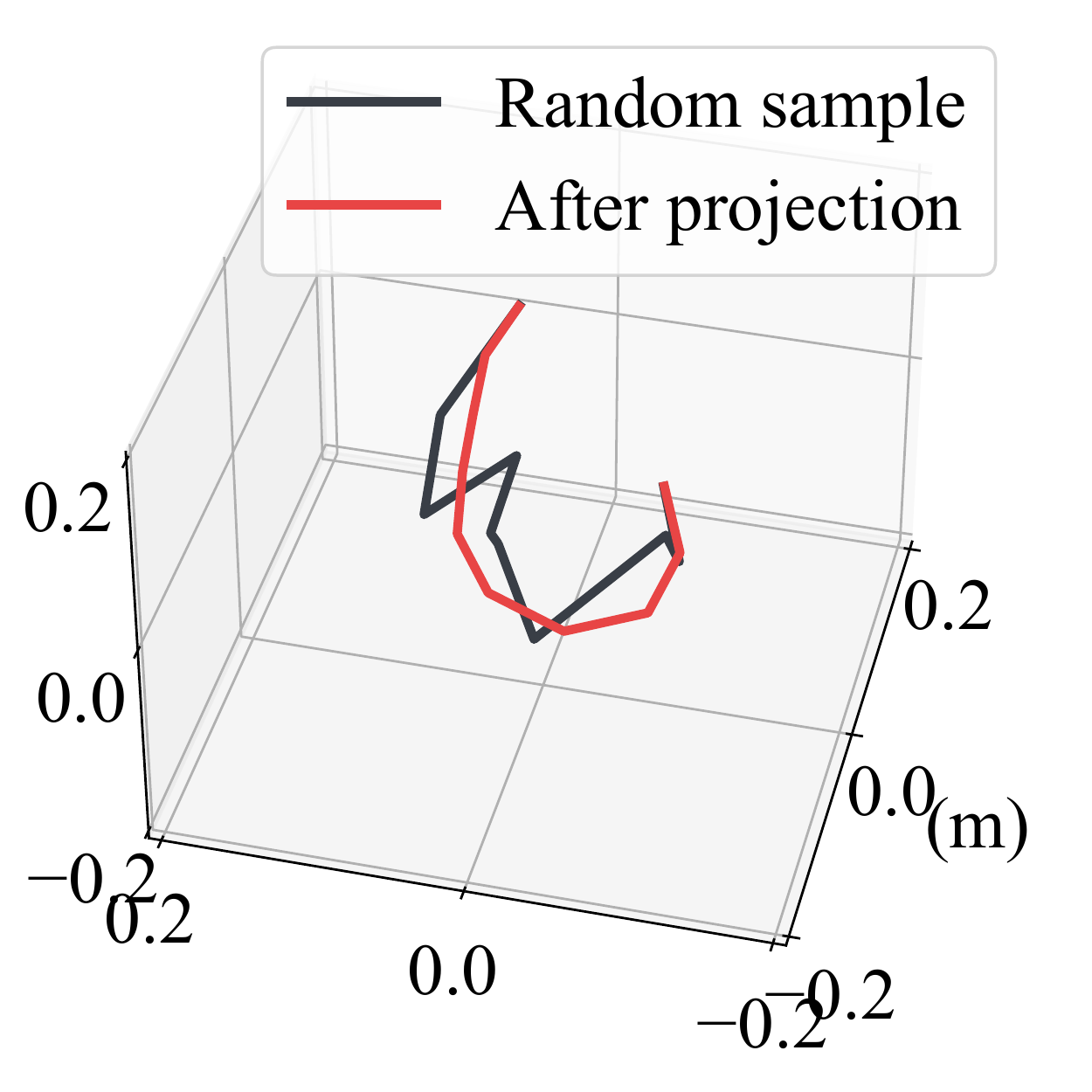} 
  }
  \hspace{-0.7cm}
  \subfigure[]{ 
    \label{fig:projection_example_2} 
    \includegraphics[width=2.5cm]{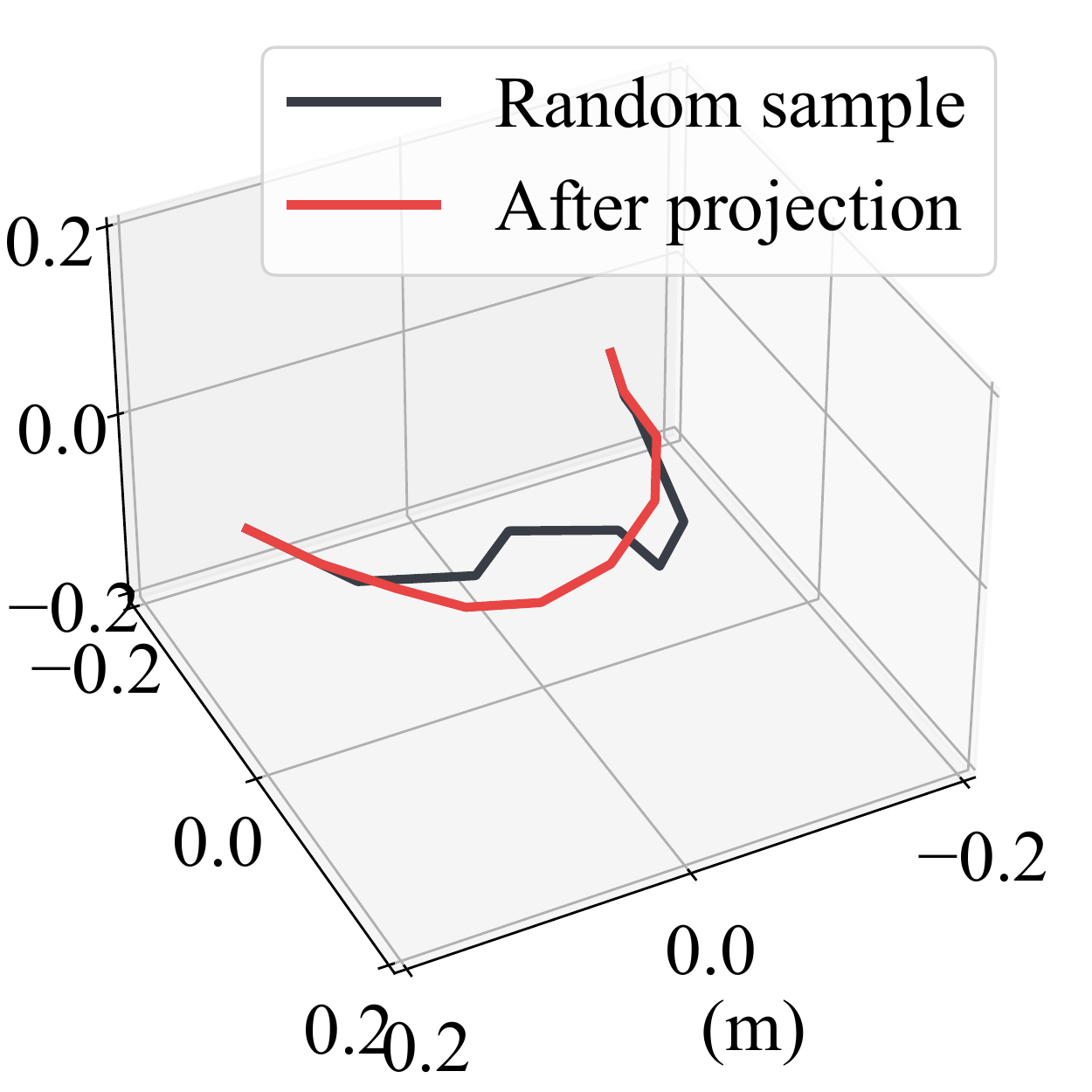} 
  }
  \vspace{-0.2cm}
  \caption{(a) Illustration of the mass-spring model which models a DLO as a series of masses connected by springs. The $i^{\rm th}$ mass is connected to $(i \pm 1)^{\rm th}$ masses by Type 1 springs and $(i \pm 2)^{\rm th}$ masses by Type 2 springs.
  (b)(c) Illustrations of projecting a randomly sampled raw DLO configuration to a stable configuration by locally minimizing the potential energy of the mass-spring model as (\ref{eq:energy_optimize}).
  }
  \vspace{-5mm}
\end{figure}

In this work, we use a simple mass-spring model \cite{yin2021modeling} as the DLO energy model (illustrated in Fig. \ref{fig:mass_spring_model}), where the energy $E$ for a configuration $\bm x$ is specified as 
\begin{equation}
\begin{aligned}
    E & = 
    \sum_{k=1}^{m-1} \frac{1}{2} \lambda_{1} \left(\|\bm x_{k+1} - \bm x_k\|_2 - \frac{L}{m-1}\right)^2
    \\
    & + \sum_{k=1}^{m-2} \frac{1}{2} \lambda_{2} \left(\|\bm x_{k+2} - \bm x_k\|_2 - \frac{2L}{m-1} \right)^2
\end{aligned}
\end{equation}
where $\lambda_1$ and $\lambda_2$ are the stiffness of Type 1 and Type 2 springs, respectively, and $L$ is the DLO length. In our simulations and experiments, we use the same stiffness values in the planning for all different DLOs. 
We will show that planning using such a simple model with imprecise parameters can provide acceptable guiding paths for the controller to achieve manipulation tasks. 
We visualize two examples of the ProjectStableConfig in Fig. \ref{fig:projection_example_1} and \ref{fig:projection_example_2}. 

\subsection{Planning Algorithm}
Our planning algorithm uses the same framework as the Constrained Bi-directional Rapidly-Exploring Random Tree (CBiRRT) method proposed in \cite{berenson2009manipulation}. The constraint projection guarantees the validity of nodes, and the bi-directional RRT significantly improves the planning efficiency. Two trees, respectively beginning from the start configuration and goal configuration, grow toward each other, until they are connected. Each node $\mathcal{N}$ in the trees contains both the DLO configuration $\bm{x}$ and the robot configuration $\bm{q}$.

During each iteration, one tree is selected as $\mathcal{T}^A$ and the other one as $\mathcal{T}^B$.
First, a random node $\mathcal{N}_{\rm rand}$ is sampled. The node in $\mathcal{T}^A$ nearest to $\mathcal{N}_{\rm rand}$ is selected as $\mathcal{N}_{\rm near}^A$. Then, $\mathcal{T}^A$ grows from $\mathcal{N}_{\rm near}^A$ towards $\mathcal{N}_{\rm rand}$ one or more steps and terminates at $\mathcal{N}_{\rm reached}^A$ using a $\text{ConstrainedExtend}$ function. 
Next, the node in $\mathcal{T}^B$ nearest to $\mathcal{N}_{\rm reached}^A$ is selected as $\mathcal{N}_{\rm near}^B$. Then,  $\mathcal{T}^B$ grows from $\mathcal{N}_{\rm near}^B$ towards $\mathcal{N}_{\rm reached}^A$ and terminates at $\mathcal{N}_{\rm reached}^B$.
If $\mathcal{N}_{\rm reached}^B$ can be connected to $\mathcal{N}_{\rm reached}^A$, then a path is found; otherwise, the two tree are swapped and next iteration starts.

Specifically, when sampling an $\mathcal{N}_{\rm rand}$, we first sample a stable DLO configuration using the $\text{ProjectStableConfig}$ function. A heuristic raw-space sampling method is preferred to sample initial configurations more like a DLO.
If the DLO is in collision, the node is discarded and a new one is generated.
In addition, the distance between two nodes is specified as the Euclidean distance between their DLO configuration vectors.

The function $\mathcal{N}_{\rm reached} = \text{ConstrainedExtend}(\mathcal{N}_{\rm from}, \mathcal{N}_{\rm to})$  iteratively moves from $\mathcal{N}_{\rm from}$ to $\mathcal{N}_{\rm to}$ and returns reached $\mathcal{N}_{\rm reached}$. 
In each step, a new DLO configuration is calculated by interpolating from the last reached configuration $\bm x_{\rm last}$ to $\bm x_{\rm to}$ with a small step size, where we use linear interpolation for the centroid positions and spherical interpolation for the relative deformations irrelevant to translations to keep the overall DLO shapes and avoid over-compression \cite{moll2006path}. The $\text{ProjectStableConfig}$ is then applied to project the interpolated configuration to a stable $\bm x_{\rm new}$.
The corresponding robot configuration $\bm q_{\rm new}$ is obtained as the inverse kinematics (IK) solution for the end poses of $\bm x_{\rm new}$, and the solution closest to $\bm q_{\rm last}$ is chosen if multiple solutions exist. If the IK succeeds and no collision occurs, a new node $\mathcal{N}_{\rm new}$ is generated. Such iteration repeats until the node generation fails or $\mathcal{N}_{\rm new}$ is farther from $\mathcal{N}_{\rm to}$ than the last reached node $\mathcal{N}_{\rm last}$.

For the start node, both the DLO and robot configurations are known; however, for the goal node, only the goal DLO configuration is given.
To use the bi-directional RRT, we set the goal robot configuration as the IK solution for the estimated DLO end poses. If multiple solutions exist, we choose the one closest to a pre-default robot configuration. Note that in actual executions, the robot with this goal configuration may not bring the DLO to the desired shape.

When the planning succeeds, the path from the start node to the goal node is extracted from the two trees and then smoothed. The final path is denoted as $\mathcal{P} = \{\mathcal{N}_{start}, \mathcal{N}_1, \cdots, \mathcal{N}_{goal}\}$ and sent to the local controller as a guiding global path.

\section{Manipulation with Local Control}
Because of the DLO modeling errors, the planned path
may not be exactly achievable. We apply an MPC to track the
planned path as guidance while using real-time feedback
to adjust the robot motion to ensure that the actual path of
both the DLO and robot is still collision-free and the final
desired DLO configuration is precisely reached.

\subsection{Manipulation Process}
Each node of the planned guiding path is regarded as an intermediate desired configuration for the controller. During the actual execution, the DLO and the manipulators are controlled to move along the path, i.e., to move to the next intermediate configuration iteratively until reaching the final desired configuration. 

Each intermediate desired configuration (except for the final desired configuration) doesn't need to be exactly reached, since it may be unachievable owing to modeling errors. If the distance between the current and desired DLO configuration is less than a threshold, the control objective will be switched to the next planned configuration. Besides, if the controller gets stuck (i.e., the control input is close to zero), the global planner will be invoked again to re-plan a path from the current configuration to the final desired configuration.

\subsection{Controlling to an Intermediate Configuration}
While moving to the next planned node $\mathcal{N}_{i}$, the desired DLO configuration $\bm x_d$ is set as $\mathcal{N}_{i}.\bm x$ and the desired robot configuration $\bm q_d$ is set as $\mathcal{N}_{i}.\bm q$.

\subsubsection{Artificial Potentials}
The attractive potential for driving the DLO to $\bm x_d$ and the attractive potential for driving the arms to $\bm q_d$ is respectively designed as
\begin{equation}
    U_a^{\rm dlo}(\bm x) = 
    \frac{1}{2} (\bm x - \bm x_d)^{\transpose} (\bm x - \bm x_d) 
\end{equation}
\begin{equation}
    U_a^{\rm arm}(\bm q) = 
    \frac{1}{2} (\bm q - \bm q_d)^{\transpose} (\bm q - \bm q_d)
\end{equation}

To achieve local obstacle avoidance of the DLO and robot, we design repulsive potentials. The repulsive potential for prevent a point $\bm p$ from colliding with obstacles $\mathcal{CO}$ in the cartesian space is designed as
\begin{equation}
    U_{r}(\bm p) = 
    \left\{ \begin{array}{cl}
    \frac{1}{2} \left( \frac{1}{c(\bm p)} - \frac{1}{\epsilon_{c}} \right)^2          & ,  \quad c(\bm p) < \epsilon_{c} \\ 
     0            & ,  \quad \rm{otherwise}
    \end{array}\right.
\end{equation}
where $c(\bm p) = \min_{\bm p' \in \mathcal{CO}} \|\bm p - \bm p'\|$ is the distance between the point and the closest obstacle, and $\epsilon_{c}$ is the maximum distance that the repulsive potential affects.
The repulsive potential for the DLO is then designed as
\begin{equation}
\begin{aligned}
     U_{r}^{\rm dlo}(\bm x) & = 
     \sum_{k=1}^{m} U_{r}(\bm x_k)
\end{aligned}
\end{equation}
Denote the Cartesian-space positions of $y$ control points on the arm bodies as $\bm \xi(\bm q) = [\bm \xi_1(\bm q); \cdots; \bm \xi_y(\bm q)]$, where $\bm \xi_i(\cdot)$ represents the forward kinematics for the $i^{\rm th}$ point. In this work we choose one control point from each arm link \cite{siciliano2009motion}.
The repulsive potential for the arms is then designed as
\begin{equation}
\begin{aligned}
    U_r^{\rm arm}(\bm q) = \sum_{i=1}^{y} U_{r}(\bm \xi_i(\bm q))
\end{aligned}
\end{equation}

\subsubsection{Control Law}
The controller is formulated as an one-step MPC. The control input is specified as the locally optimal solution of the following optimization problem:
\begin{equation} \label{eq:control_law}
\begin{aligned}
    \min_{\bm u} \, \quad &
    \mathcal{J} = 
      \lambda_a^{\dlo} \, U_{a}^{\dlo}(\bm x(t+\delta t))
    + \lambda_r^{\dlo} \, U_{r}^{\dlo}(\bm x(t+\delta t))
    \\
    & \quad 
    + \lambda_a^{\arm} \, U_{a}^{\arm}(\bm q(t+\delta t))
    + \lambda_r^{\arm} \, U_{r}^{\arm}(\bm q(t+\delta t))
    \\
    & \quad 
    + \frac{1}{2} \bm u^\transpose \bm K_u \bm u
    \\
    \text{s.t.} 
    \quad 
        & \bm C_{\rm dof} \bm J^{\arm} \bm u = \bm 0
    \\
        &  \bm u^\transpose \bm K_u \bm u \leq u_{\rm max}^2
\end{aligned}
\end{equation}
where $\lambda_a^{\dlo}, \lambda_r^{\dlo}, \lambda_a^{\arm}, \lambda_r^{\arm}$ are weighting coefficients for the potentials, and $\delta t$ is the step interval. The $\bm K_u$ is a weighting matrix for the input joint velocities, and related terms are for penalizing and constraining the input magnitude. 
The first equality constraint is to constrain the allowed DoFs of the robot end-effectors for specific tasks (e.g., 2-D tasks), where $\bm J^{\arm}$ is the robot Jacobian matrix. 

To estimate the DLO configuration at the next time step given the control input $\bm u$, we use a DLO Jacobian matrix which relates the velocity of DLO features to the velocity of robot end-effectors as $\dot{\bm x} = \bm J^{\dlo} \dot{\bm r}$.
We employ our previously proposed offline-online data-driven method \cite{yu2022global} to obtain an estimation $\hat{\bm J}^{\dlo}$. In addition, $\dot{\bm r} = \bm J^{\arm} \dot{\bm q}$.
Then, the configuration at the next step is approximated as 
\begin{equation}
    \bm q(t+\delta t)  \approx  \bm q(t) + \bm u \delta t
\end{equation}
\begin{equation}
    \bm x(t+\delta t)  \approx  \bm x(t) + \hat{\bm J}^{\dlo} \bm J^{\arm} \bm u \delta t
\end{equation}

To avoid the huge computation of incorporating the distance calculation into each optimization iteration, we linearize the repulsive potential terms in the cost function as
\begin{equation}
    U_{r}^{\arm}(\bm q(t+\delta t))
     \approx
     U_{r}^{\arm}(\bm q(t))
     +
    \nabla_{\bm q} U_r^{\arm}(\bm q(t)) 
    \bm u \delta t
\end{equation}
\begin{equation}
    U_{r}^{\dlo}(\bm x(t+\delta t))
     \approx
    U_{r}^{\dlo}(\bm x(t))
    +
    \nabla_{\bm x} U_r^{\dlo}(\bm x(t)) 
    \hat{\bm J}^{\dlo} \bm J^{\arm} \bm u \delta t
\end{equation}
Then, only the signed distances at the current time $t$ are required, and the MPC can be efficiently solved using gradient-based optimization solvers. 

When controlling to the final desired configuration, there is no need to consider the planned arm configuration, so $\lambda_a^{\arm}$ is set to zero.

\section{Results}

In both the simulations and experiments, the DLO is grasped by a dual-UR5 robot and its configuration is represented by 10 features ($m = 10$). The control frequency is 10 Hz.
The stiffness of the mass-spring model is set as $\lambda_1 = 1.0 $ and $\lambda_2 = 1.0$. 
Other hyper-parameters are set as 
$\epsilon_c = 0.1$, 
$\lambda_a^{\dlo} = 10.0$, 
$\lambda_a^{\arm} = 1.0$, and 
$\lambda_r^{\dlo} = \lambda_r^{\arm} = 10^{-3}$.
The planning algorithm is written in C++ and runs on an Intel i7-10700 CPU (2.9GHz), where the optimization is based on the Ceres Solver \cite{Agarwal_Ceres_Solver_2022}.

We define the following criteria for evaluation:
1) planning success: the planner finds a feasible path within 1000 RRT iterations;
2) planning time: the time cost for finding a feasible path;
3) final task error: the Euclidean distance between the final desired DLO configuration and reached configuration in a manipulation;
4) collision time: the time (second) of collisions between the DLO, robot, and obstacles during a manipulation;
5) manipulation success: the final task error is less than 5cm.
Note that the task error contains errors of all features on the DLO without averaging.

\begin{figure} [tb]
  \centering 
    \includegraphics[width=8.6cm]{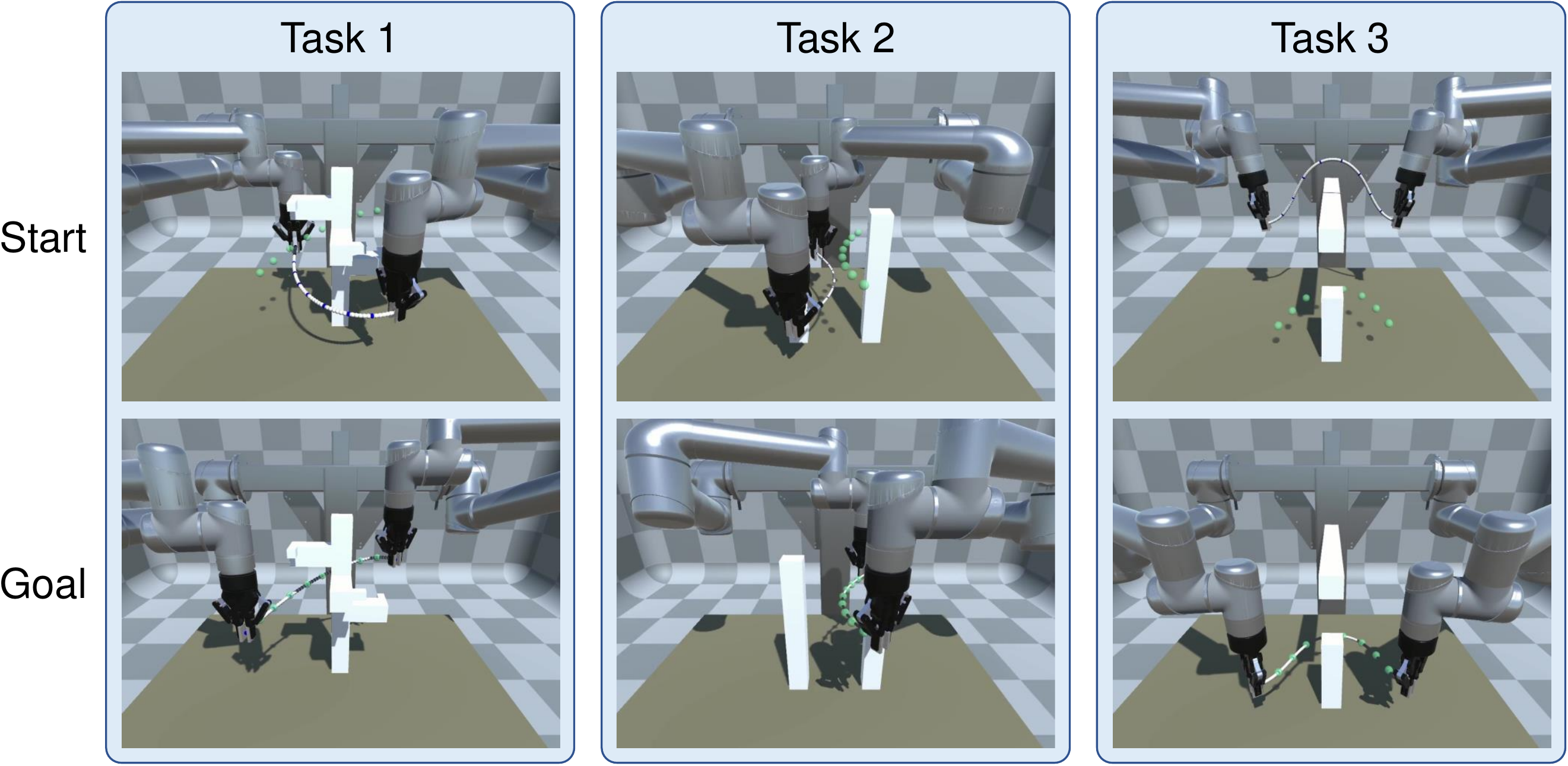} 
  \vspace{-5mm}
  \caption{Three simulated tasks for testing the overall performance of the proposed method. The top row is the start configurations and the bottom row is the goal configurations. The pictures are taken from actual manipulations.}
  \label{fig:exp_sim_3tasks}
\end{figure}

\begin{table}
\centering
\caption{Performance of our method in the three simulated tasks.}
\label{tab:exp_sim_overall_performance}
\begin{threeparttable}[b]
\begin{tabular}{c|cccc} 
\toprule
Task & \begin{tabular}[c]{@{}c@{}}Planning\\success rate~\end{tabular} & \begin{tabular}[c]{@{}c@{}}Planning\\time (s)\tnote{a}\end{tabular} & \begin{tabular}[c]{@{}c@{}}Manipulation\\success rate\end{tabular} & \begin{tabular}[c]{@{}c@{}}Final task\\error (cm)\tnote{b}\end{tabular} \\ 
\hline
1 & 8/10 & 3.29~$\pm$~1.45 & 8/8 & 0.55~$\pm$~0.22 \\
2 & 8/10 & 4.14~$\pm$ 4.31 & 8/8 & 0.55~$\pm$~0.11 \\
3 & 10/10 & 2.66~$\pm$~1.25 & 10/10 & 0.60 $\pm$~0.18 \\
\bottomrule
\end{tabular}
\begin{tablenotes}
     \item[a,b] The mean value $\pm$ standard deviation over the successful cases.
\end{tablenotes}
\end{threeparttable}
    \vspace{-3mm}
\end{table}

\subsection{Simulations}
The simulation is built in Unity \cite{unity} with Obi \cite{obi} for simulating DLOs and Unity Robotics Hub \cite{unity_robotics_hub} for integration with ROS. The simulator is utilized as a black box.

\subsubsection{Overall Performance}
We first test the overall performance of our method in three different tasks with different scenarios, DLO properties, and start/goal configuration, as shown in Fig \ref{fig:exp_sim_3tasks}. The lengths of the DLOs are 0.5/0.4/0.6m for Task 1/2/3, respectively. 
For each task, the planner tries 10 queries and all found paths are executed by the controller. 
The performance is summarized in Table \ref{tab:exp_sim_overall_performance}. The planning success rate over all queries is 87\%, and most of the successful queries take about 2\textasciitilde5 seconds to find a feasible path. All 26 paths are successfully executed by the controller, and the final tasks are achieved with an average final task error of 0.57cm. Re-planning is invoked 2 times during all 26 manipulations. One ProjectStableConfig takes 0.7ms on average.
It is demonstrated that our planning and control framework can efficiently and robustly achieve various DLO manipulation tasks.

\begin{figure} [tb]
  \centering 
    \includegraphics[width=8.6cm]{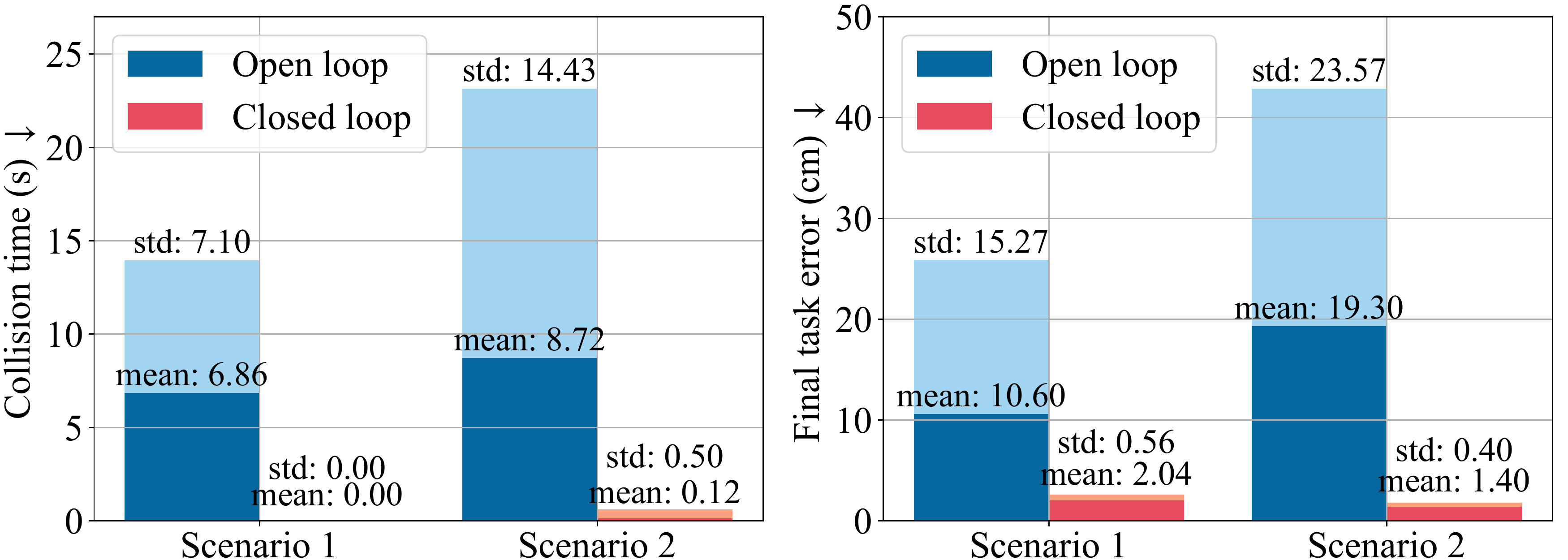} 
  \vspace{-5mm}
  \caption{Quantitative results of the comparison between the proposed closed-loop framework and the open-loop manner. The bars with darker colors refer to the mean values over all 25 manipulations, and those with lighter colors refer to the standard deviations (std).}
  \label{fig:exp_sim_w_o_control_compare}
    \vspace{-3mm}
\end{figure}

\begin{figure*} [tb]
  \centering 
  \subfigure[]{ 
    \label{fig:exp_sim_two_cases_1} 
    \includegraphics[width=0.48\textwidth]{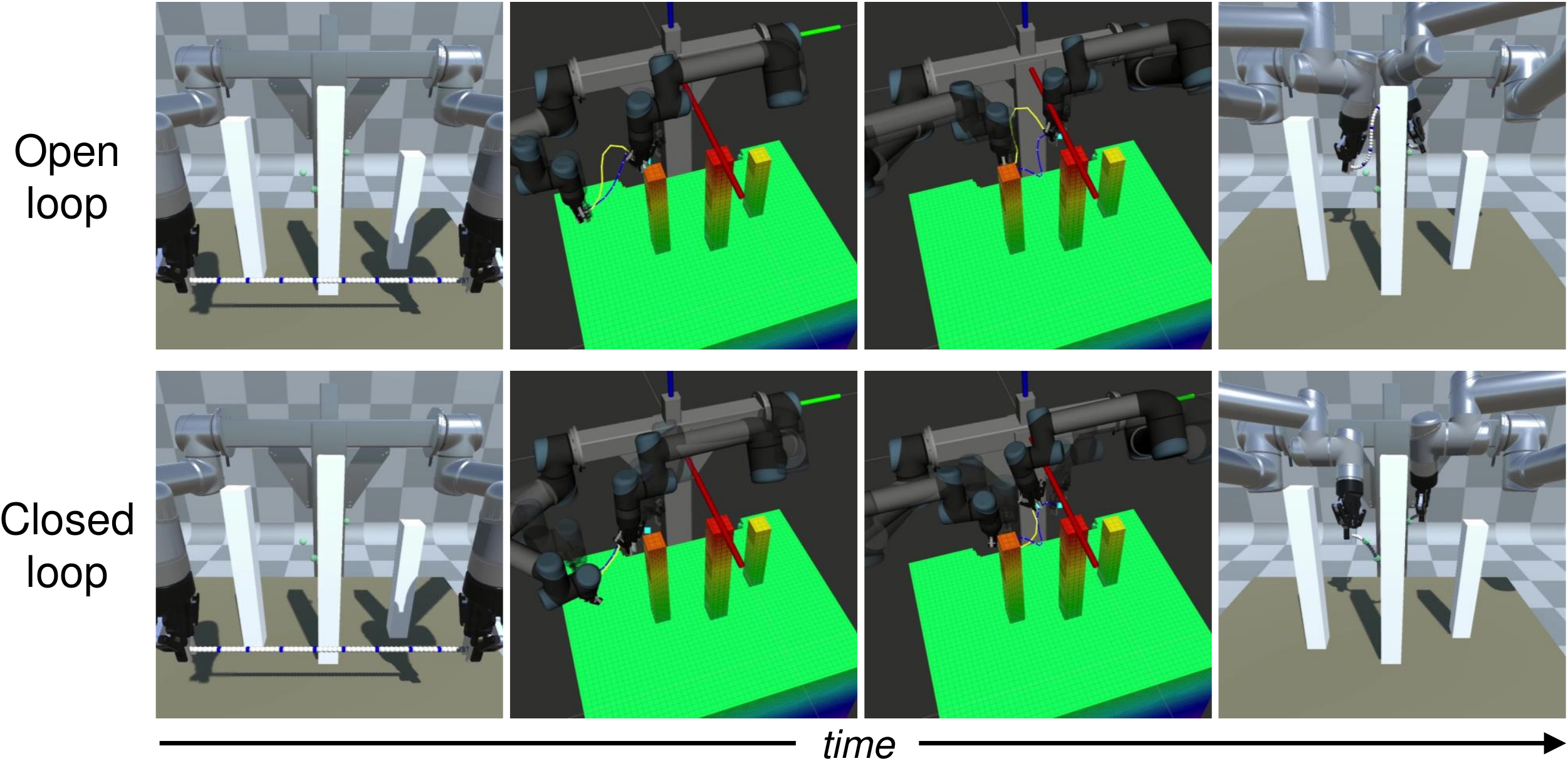} 
  } 
  \subfigure[]{ 
    \label{fig:exp_sim_two_cases_2} 
    \includegraphics[width=0.48\textwidth]{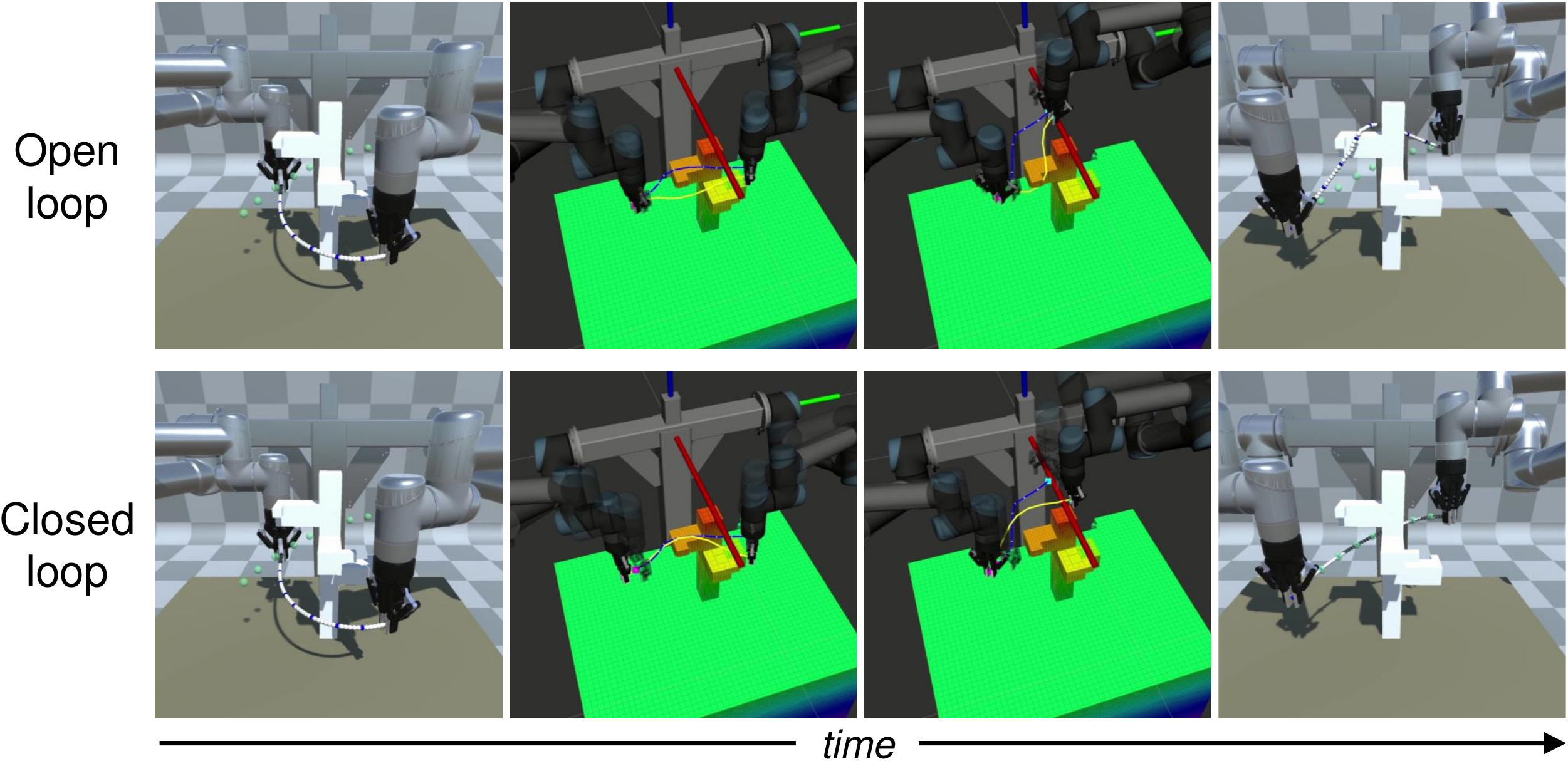} 
  }
  \vspace{-0.3cm}
  \caption{Two cases for illustrating the comparison between the open-loop manner and proposed closed-loop manner using the same planned paths (more clearly shown in the supplementary video).
  The first column is the start configurations; the second and third columns are the manipulation processes, where the blue lines and translucent robots refer to the planned intermediate configurations and the yellow lines and non-translucent robots refer to the real-time configurations; and the last column is the final reached configurations, where the translucent green points refer to the final desired DLO configurations.}
  \label{fig:exp_sim_two_cases}
\end{figure*}

\begin{figure*} [tb]
  \centering 
    \includegraphics[width=0.8\textwidth]{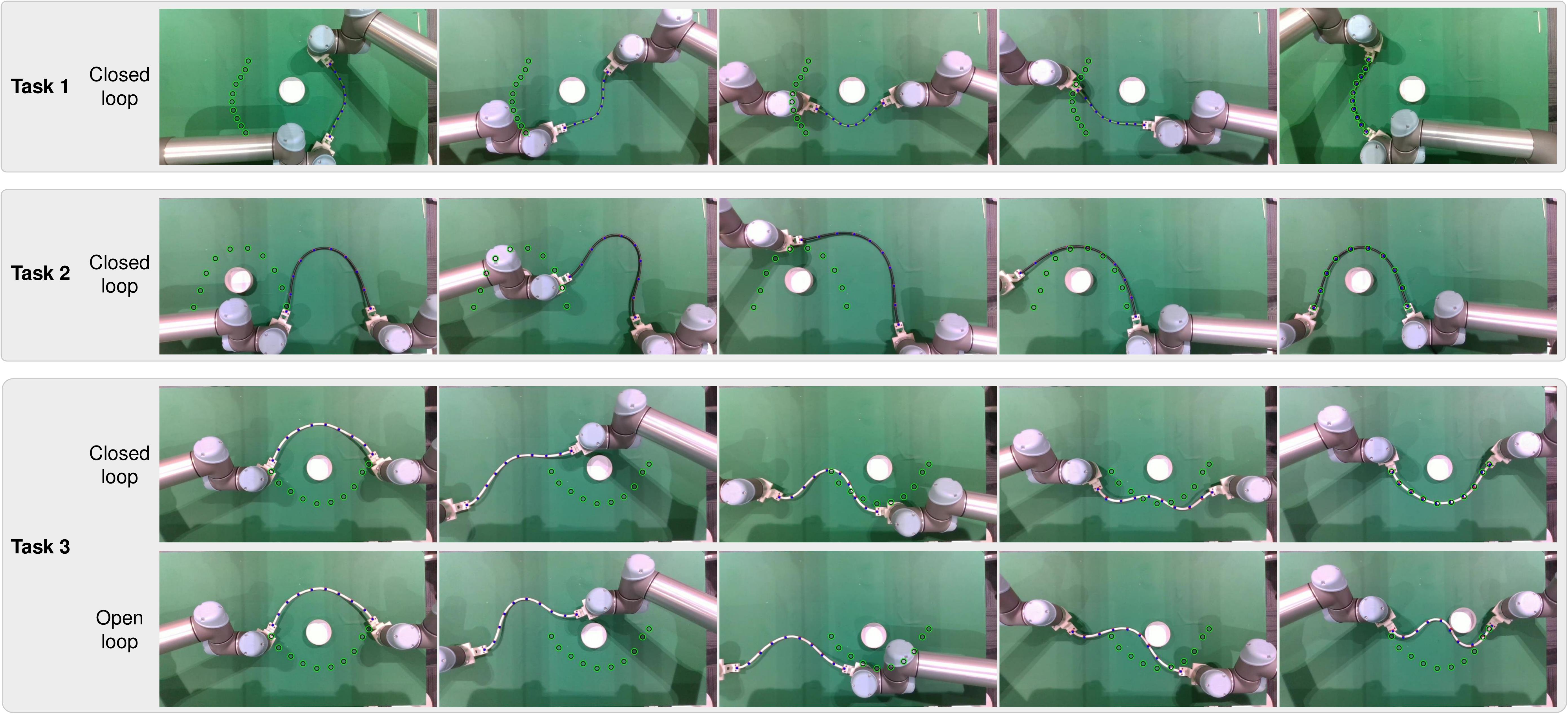} 
  \vspace{-2mm}
  \caption{Manipulation processes of three real-world tasks using the proposed planning and control framework. The green+black circles refer to the final desired DLO configurations. For Task 3, we also show the comparison between the proposed closed-loop manner and the open-loop manner.}
  \label{fig:exp_real}
    \vspace{-2mm}
\end{figure*}

\subsubsection{Closed-Loop v.s. Open-Loop}

We verify the significance of the control module by comparing the actual manipulation results of directly executing the planned robot path in an open-loop manner and using the proposed controller to track the planned guiding path in a closed-loop manner.

We both do quantitative tests to show the statistical comparison results and visualize two specific cases which clearly illustrate the differences. 
The quantitative tests use the same two scenarios as the two cases in Fig. \ref{fig:exp_sim_two_cases}, and the DLO lengths are 0.4/0.6m, respectively. The DLOs start from straight lines in front of the obstacles. Five different desired shapes behind the obstacles are tested for each scenario, and five paths are planned for each desired shape, so totally 25 paths are executed in both open-loop and closed-loop manners for each scenario. Fig. \ref{fig:exp_sim_w_o_control_compare} shows the results of the collision time during manipulations and the final task errors. It can be seen that 1) the proposed closed-loop manner can robustly accomplish DLO manipulation tasks using coarse planned paths and achieve precise final control results (1.77cm on average), while the open-loop manner cannot guarantee that the DLO moves as expected like the case in Fig. \ref{fig:exp_sim_two_cases_1}; 2) the proposed closed-loop manner can 
effectively avoid collisions (0.06s on average), while collisions often happen in the open-loop manipulations, which may cause failures of tasks like the case in Fig. \ref{fig:exp_sim_two_cases_2}.

For better illustration, we visualize the manipulation processes of two specific cases in Fig. \ref{fig:exp_sim_two_cases}. 
As for the first case shown in Fig. \ref{fig:exp_sim_two_cases_1}, in the open-loop manipulation, although the robot moves along the planned robot path, the DLO gradually deviates from the planned DLO path and finally terminates at another stable but undesired shape; while in the closed-loop manipulation, the controller receives the real-time DLO configuration feedback and adjusts the robot motion, 
during which the actual robot path is not exactly as planned. As for the second case shown in Fig \ref{fig:exp_sim_two_cases_2}, in the open-loop manipulation, unexpected collisions between the DLO and obstacles happen and the DLO is hooked, resulting in the failure of the task; while in the closed-loop manipulation, thanks to the local obstacle avoidance achieved by the controller using artificial potentials, the DLO avoids the obstacles and finally reaches the desired configuration.

\subsection{Real-World Experiments}

We also validate the proposed framework in the real world. 
To simplify the perception of DLOs, we put markers along the DLOs and conduct 2-D tasks where the DLOs are placed on a table to avoid visual occlusion. 

Three representative tasks are achieved using the proposed method, as shown in Fig. \ref{fig:exp_real}. In Task 1, the DLO needs to be rotated around 180 degrees; in Task 2, the method in \cite{mcconachie2020manipulating} may fail, since the "virtual elastic band" connecting the two end-effectors will not collide with the obstacle along a straight-line robot path, but the actual DLO body will; and in Task 3, the DLO needs to be manipulated from an upper-semicircle shape to a lower-semicircle shape.
A 0.3m-long nylon rope, a 0.6m-long HDMI cable, and a 0.45m-long electric wire are used in Task 1, 2, and 3, respectively. 
For Task 3, we also show the comparison between the proposed closed-loop manner and the open-loop manner using the same planned path, in which the open-loop execution moves the DLO to an unexpected shape and also causes the collision, while the closed-loop execution succeeds.




\section{Conclusion}

This paper proposes a coarse-to-fine framework for dual-arm manipulation of DLOs with whole-body obstacle avoidance.
A coarse global path is first efficiently planned using an RRT planning framework, in which a DLO energy model is employed to generate stable DLO configurations. 
Then, an MPC controller tracks the planned path as global guidance, during which it reacts to the real-time states in a closed-loop manner to compensate for the modeling errors and avoid collisions. 
This formulation considers both the feasibility and accuracy of long-horizon manipulation tasks in constrained environments.
The results show that our method can robustly and precisely achieve various DLO manipulation tasks in both simulations and real-world experiments, where the necessity of such a closed-loop framework is also demonstrated by comparative studies.

Future work will be devoted to 
incorporating more accurate discrete DLO energy models and adding gravity energy terms, further tuning each component of the planning algorithm, employing long-horizon MPC for local control, or applying marker-free perception methods \cite{lv2022learning}.







\addtolength{\textheight}{-9.0cm}   

\clearpage

\bibliographystyle{IEEEtran}
\bibliography{ref}

\clearpage

\end{document}